\documentclass[11pt]{article}
\usepackage{coling2020}
\usepackage{times}
\usepackage{url}
\usepackage{latexsym}
\usepackage{multirow}
\usepackage{graphicx}
\usepackage{tabularx}
\usepackage{booktabs}
\usepackage{amsmath}
\usepackage{adjustbox}

\usepackage[linesnumbered,ruled]{algorithm2e}

\SetKwInput{KwInput}{Input}
\SetKwInput{KwOutput}{Output}

\newcommand{\ACRO}[1]{\textsc{#1}}
\newcommand{\ARABERT}{\ACRO{AraBERT}}
\newcommand{\BERT}{\ACRO{bert}}
\newcommand{\CONLL}{\ACRO{conll}}
\newcommand{\ELMO}{\ACRO{ELMo}}
\newcommand{\NLP}{\ACRO{nlp}}

\colingfinalcopy 

\title{Neural  Coreference Resolution for Arabic}

\author{Abdulrahman Aloraini$^{1, 2}$*\\
  \\
 \\\And
 Juntao Yu$^{1}$*\\
  $^1$Queen Mary University of London, United Kingdom \\ 
  $^2$Qassim University, Saudi Arabia \\ 
  {\tt \{a.aloraini, juntao.yu, m.poesio\}@qmul.ac.uk}
  \\\And
  Massimo Poesio$^{1}$ \\
    \\
    \\
  }

\date{}

\begin{document}
\maketitle
\begin{abstract}
 No neural coreference resolver for Arabic exists, in fact we are not aware of any learning-based coreference resolver for Arabic since \cite{anders:2014}.
 In this paper, we introduce a coreference resolution system for Arabic based on Lee et al's end-to-end architecture combined with the Arabic version of {\BERT} and an external mention detector.
 As far as we know, this is the first neural  coreference resolution system aimed specifically to Arabic, and it substantially outperforms the existing state-of-the-art on  OntoNotes 5.0 with a gain of 15.2 points {\CONLL} F1. 
 We also discuss the current limitations of the task for Arabic and possible approaches that can tackle these challenges.
 \blfootnote{
    \hspace{-0.65cm}  
    This work is licensed under a Creative Commons
    Attribution 4.0 International License.
    License details:
    \url{http://creativecommons.org/licenses/by/4.0/}.
} \let\thefootnote\relax\footnotetext{* Equal contribution. Listed by alphabetical order.} 
\end{abstract}

\section{Introduction}
\label{sec:introduction}
Coreference resolution is the task of grouping mentions in a text that refer to the same real-world entity into clusters \cite{poesio-stuckardt-versley:book} .
Coreference resolution is a difficult task 
that 
requires reasoning, context understanding, and background knowledge of real-world entities, 
and
has driven research in both natural language processing and machine learning, particularly since the release of the \textsc{ontonotes} multilingual corpus providing annotated coreference data for Arabic, Chinese and English and used for the 2011 and 2012 {\CONLL} shared tasks \cite{conll:2012}.
Since then, 
there has been substantial research on English coreference,
most recently using neural coreference approaches \cite{lee:2017,lee:2018,kantor2019bertee,joshi2019bert,joshi2019spanbert,yu-etal-2020-cluster,wu-etal-2020-corefqa}, leading to a significant increase in 
the  performance of coreference resolvers for English.
%
By contrast, there has been almost no research on Arabic coreference; 
the performance for Arabic coreference resolution has not improved  much since the {\CONLL} 2012 shared task,
and in particular no neural architectures have been proposed--the current state-of-the-art system remains 
the model proposed in 
\cite{anders:2014}. 
In this paper we close this very obvious gap by proposing what to our knowledge is the first neural coreference resolver for Arabic.\footnote{The code is available at \url{https://github.com/juntaoy/aracoref}}

\begin{figure}[h]
  \includegraphics[scale=0.6]{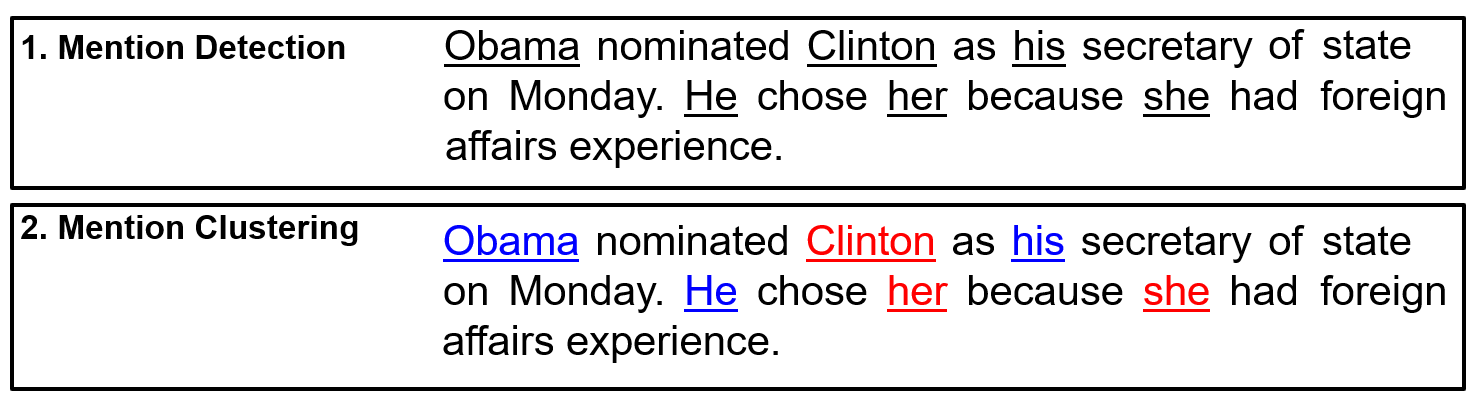}
  \centering
  \caption{The first step in coreference resolution is mention detection. The detected mentions are underlined. The second step is mention clustering. We have two clusters \{Obama, his, he\} and  \{Clinton, her, she\}. The mention detector might identify other words as mentions, but for simplicity we present only the mentions of the two clusters.}
  \label{fig:example0}
\end{figure}

One explanation for this lack of research
might simply be the lack of  training data large enough for the task. 
Another explanation 
might be that Arabic is 
more problematic than English 
because of its rich morphology, 
its 
many dialects, 
and/or 
its 
high degree of ambiguity. 
We explore the first of these possibilities. 
Coreference resolution can be further divided into two subtasks--mention detection and mention clustering--as illustrated in 
Figure \ref{fig:example0}. 
%
In  early work, coreference's two subtasks were usually carried out in a pipeline fashion \cite{soon-et-al:CL01,fernandes:2014,anders:2014,wiseman2015learning,wiseman2016learning,clark2016deep,clark2016improving}, with candidate mentions selected prior the mention clustering step. 
Since 
\newcite{lee:2017} introduced 
an end-to-end neural coreference architecture 
that achieved state of the art 
by carrying out the two tasks jointly, as first proposed by \newcite{daume&marcu:NAACL05},
most state-of-the-art systems have followed this approach.
However, no end-to-end solution was attempted for Arabic. We intend to explore whether an end-to-end solution would be practicable with a corpus of more limited size.


The approach we followed to adapt
the
state-of-the-art English coreference resolution architecture 
to Arabic 
is as follows.
We started with a strong baseline system \cite{lee:2018,kantor2019bertee},
enhanced  with contextual {\BERT} embeddings \cite{bert:2019}.
We then explored three methods for improving the model's performance for  Arabic. 
The first method is to pre-process  Arabic words with heuristic rules. 
We follow \newcite{althobaiti:2014a} to normalize the letters with different forms, and removing all the diacritics.
This results in a substantial improvement of 7 percentage points over our baseline. 
The second route is to replace  multilingual {\BERT} with a {\BERT} model trained only on the Arabic texts (\ARABERT) \cite{Antoun:2020}. 
Multilingual {\BERT} is trained with 100+ languages; as a result, 
it is not optimized for any of them.
As shown by \newcite{Antoun:2020}, monolingual {\BERT}  trained only on the Arabic texts has better performance on various {\NLP} tasks.  
We found the same holds for coreference:
using embeddings from  monolingual {\BERT}, the model further improved the {\CONLL} F1 by 4.8 percentage points. Our third step is to leverage the end-to-end system with a separately trained mention detector \cite{yu-etal-2020-neural}. 
We show that a better mention detection performance can be achieved by using a separately trained mention detector. 
And by using a hybrid training strategy between the end-to-end and pipeline approaches (end-to-end annealing) our system gains an additional 0.8 percentage points. 
Our final system achieved a {\CONLL} F1 score of 63.9\%, which is is 15\% more than the previous state-of-the-art system \cite{anders:2014} on Arabic coreference with the {\CONLL} dataset. 
Overall, we show that the state-of-the-art English coreference model can be adapted to Arabic coreference  leading to a substantial improvement in performance when compared to previous feature-based systems. 

\section{Related Work}
\label{sec:related_work}

\subsection{English Coreference Resolution}
Like with other natural language processing tasks, most state-of-the-art coreference resolution systems are evaluated  on English data. 
Coreference resolution for English is an active area of research. 
Until the appearance of neural systems, 
state-of-the-art systems for English coreference resolution  
were either 
rule-based \cite{lee:2011} or feature-based \cite{soon-et-al:CL01,bjorkelund:2011,fernandes:2014,anders:2014,clark2015entity}. 
\newcite{wiseman2015learning} introduced a neural network-based approach to solving the task in a non-linear way. 
In their system, the heuristic features 
commonly used in linear models are transformed by a $\tanh$ function to be used as the mention representations. 
\newcite{clark2016improving} integrated reinforcement learning to let the model optimize directly on the B$^3$ scores. 
\newcite{lee:2017} first presented a neural joint approach for mention detection and coreference resolution. Their model does not rely on parse trees; instead, the system learns to detect mentions by exploring the outputs of a bi-directional LSTM.  
\newcite{lee:2018}
is an extended version of \newcite{lee:2017} mainly enhanced by using {\ELMO} embeddings \cite{peters2018elmo}, 
in addition, the use of second-order inference enabled the system to explore partial entity level features and further improved the system by 0.4 percentage points. 
Later the model was further improved by \newcite{kantor2019bertee} who use  {\BERT} embeddings \cite{bert:2019} instead of {\ELMO} embeddings. 
In these systems, 
both {\BERT} and {\ELMO} embeddings are used in a pre-trained fashion. 
More recently, \newcite{joshi2019bert} fine-tuned the {\BERT} model for  coreference, resulting in a small further improvement. 
Later, \newcite{joshi2019spanbert} introduces \ACRO{SpanBERT} which is trained for  tasks that involve spans.
Using  \ACRO{SpanBERT}, 
they 
achieved a substantial gain of 2.7\% when compared with the \newcite{joshi2019bert} model. \newcite{wu-etal-2020-corefqa} reformulate the coreference resolution task as question answering task and achieved the state-of-the-art results by pretrain the system first on the large question answering corpora. 

\subsection{Arabic Coreference Resolution}

There have been several studies of Arabic coreference resolution; in particular, several of the systems involved in the {\CONLL} 2012 shared task attempted Arabic as well. 
\newcite{li:2012} used syntactic parse trees to detect mentions, and compared pairs of mention based on their semantic and syntactic features. \newcite{zhekova:2010} proposed a language independent module that requires only  syntactic information and clusters mentions using the memory-based learner TiMBL \cite{daelemans:2004}. 
\newcite{chen:2012} detected mentions by employing named entity and language-dependent heuristics. 
They employed multiple sieves \cite{lee:2011} for English and Chinese, but  only used an exact match sieve for Arabic because other sieves did not provide better results. 
\newcite{bjorkelund:2011} considered all noun phrases and possessive pronouns as mentions, and trained two types of classifier: logistic regression and decision trees. \newcite{stamborg:2012} extracted all noun phrases, pronouns, and possessive pronouns as mentions. Then they applied \cite{bjorkelund:2011}'s solver which consists of various lexical and graph dependency features. 
\newcite{uryupina:2012} adapted  for Arabic the BART \cite{Versley:2008} coreference resolution system, which consists of five components:  pre-processing pipeline, mention factory, feature extraction module, decoder and encoder. 
\newcite{fernandes:2014} defined a set of rules  based on parse tree information to detect mentions, and utilized a latent tree representation to learn coreference chains. 
Similarly \newcite{anders:2014} adopted a tree representation approach to cluster mentions, but improved the learning strategy and introduced non-local features to capture more information about coreference relations. 
There have been other research studies related to anaphora resolution \cite{trabelsi2016arabic,bouzid2017combine,beseiso2016coreference,abolohom2015hybrid}, but they only considered  pronominal anaphora. \newcite{aloraini2020cross} also considered a specific type of pronominal anaphora,  zero-pronoun anaphora. 
All current approaches suffer from a number of limitations, one of which is that most of them rely on an extensive set of hand-chosen features.

\section{System architecture}
\label{sec:model}
\subsection{The Baseline System}
\label{sec:baseline}
We use the
\newcite{lee:2018} system as our baseline and replace their {\ELMO} embeddings with the {\BERT} recipe of \newcite{kantor2019bertee}. 
The input of the system is the concatenated embeddings ($(emb_t)_{t=1}^{T}$) of both word and character levels.
The word-level fastText \cite{bojanowski2016enriching} and {\BERT} \cite{bert:2019} embeddings are used together with the character embeddings learned from a convolution neural network (CNN) during training. The input is then put through a multi-layer bi-directional LSTM to create the token representations ($(x_t)_{t=1}^{T}$).
The $(x_t)_{t=1}^{T}$  are used together with head representations ($h_i$) to form the mention representations ($M_i$). 
The $h_i$ of a mention is calculated as the weighted average of its token representations ($\{x_{b_i}, ..., x_{e_i}\}$), where $b_i$ and $e_i$ are the indices of the start and the end of the mention respectively. The mention score ($s_m(i)$) is then computed by a feedforward neural network to determine the likeness of a candidate to be mention.
Formally, the system computes $h_i$, $M_i$ and $s_m(i)$ as follows:

$$\alpha_t = \textsc{ffnn}_{\alpha}(x_{t})$$
$$a_{i,t} = \frac{exp(\alpha_t)}{\sum^{e_i}_{k=b_i} exp(\alpha_k)}$$
$$h_i = \sum^{e_i}_{t=b_i} a_{i,t} \cdot x_t$$
$$M_i = [x_{b_i}, x_{e_i},h_i,\phi(i)]$$
$$s_m(i) = \textsc{ffnn}_m(M_i)$$
where $\phi(i)$ is the mention width feature embeddings. 
To make the task computationally tractable, the system only considers mentions up to a maximum width of 30 tokens (i.e. $e_i - b_i < 30$). Further pruning on candidate mentions is applied before approaching the antecedent selection step. The model keeps a small portion (0.4 mention/token) of the top-ranked spans according to their mention scores ($s_m(i)$).

Next, the system uses a bilinear function to compute a light-weight mention pair scores ($s_c(i,j)$) between all the valid mention pairs\footnote{Candidate mentions are paired with all the mentions appeared before them (candidate antecedents) in the document.}. The scores are then used to select top candidate antecedents for all candidate mentions (coarse antecedent selection). More precisely, the $s_c(i,j)$ are computed as follows:

$$s_c(i,j) = M_{i}^{\top} W_c M_{j}$$

After that, the system further computes a more accurate mention pair scores between the mentions and their top candidate antecedents $s_a(i,j)$:  

$$P_{(i,j)} = [M_i, M_j, M_i \circ M_j,\phi(i,j)]$$
$$s_a(i,j) = \textsc{ffnn}_a(P_{(i,j)})$$
where $P_{(i,j)}$ is the mention pair representation, $M_i$, $M_j$ is the representation of the antecedent and anaphor respectively, 
$\circ$ denotes  element-wise product, and
$\phi(i,j)$ is the distance feature between a mention pair.

The next step is to compute the final pairwise score ($s(i,j)$). The system adds an artificial antecedent $\epsilon$ to deal with cases of non-mentions, discourse-new mentions or cases when the antecedent does not appear in the candidate list. The $s(i,j)$ is calculated as follows:

$$
s(i,j) = \Bigg\{
  \begin{tabular}{ll}
  0& $i=\epsilon$ \\
  $s_m(i)+s_m(j)+s_c(i,j)+s_a(i,j)$& $i\neq \epsilon$
  \end{tabular}
$$

For each mention the predicted antecedent is the one that has the highest $s(i,j)$. An anaphora-antecedent link will be created only if the predicted antecedent is not $\epsilon$.

\begin{figure}[t]
    \centering
    \includegraphics[width=\columnwidth]{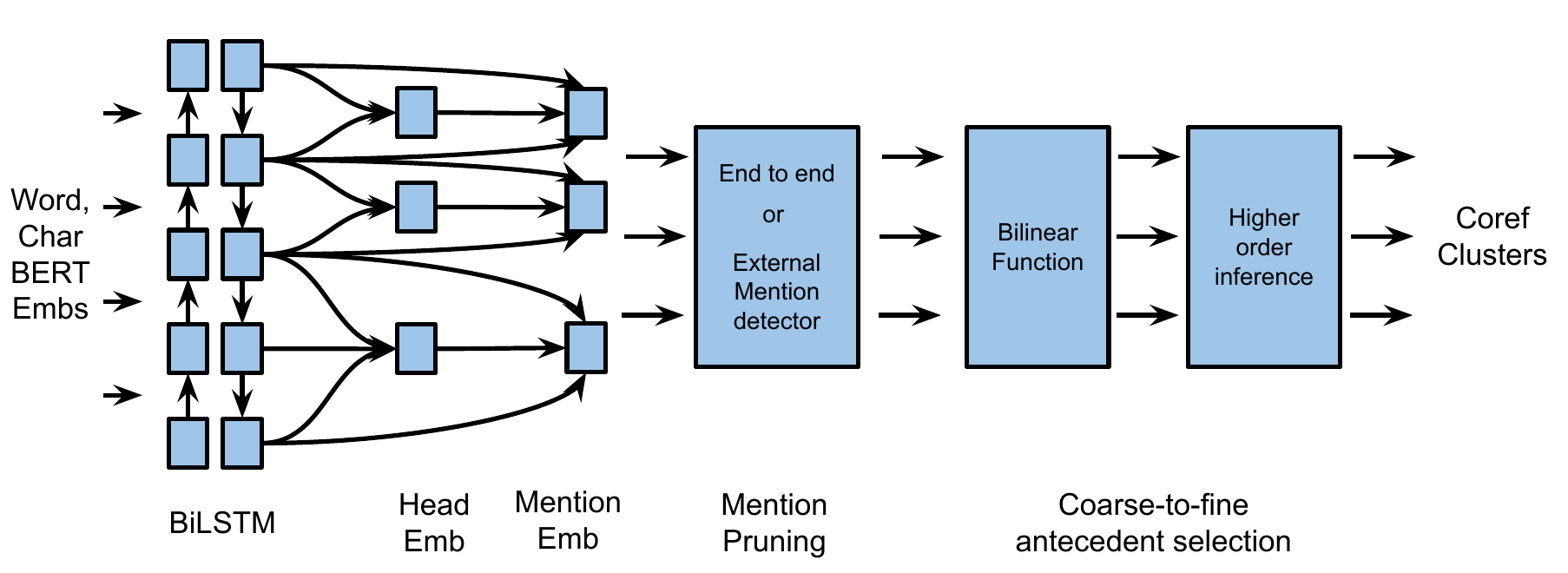}
    \caption{The proposed system architecture.
    }
    \label{fig:nngraph}
\end{figure}

Additionally, the model has an option to use higher-order inference to allow the system to access entity level information. We refer the reader to the original \newcite{lee:2018} paper for more details. We use the default setting of \newcite{lee:2018} to do second-order inference. The final clusters are created using the anaphora-antecedent pairs predicted by the system. Figure \ref{fig:nngraph} shows the proposed system architecture of our system.

\subsection{Data Pre-processing}

Arabic is a morphologically rich language.
Thus, training on Arabic texts that are not  pre-processed properly can suffer from sparsity (various forms for the same word) and ambiguity (same form corresponding to multiple words). 
There are two reasons for these problems. First, certain letters can have different forms which are usually misspelled, such as the various forms of the letter ``alif''. Second, the placement of diacritics on words which are assumed to be undiacritized \cite{habash2006arabic}.  
Therefore, we follow the  steps proposed in \cite{althobaiti:2014a} to pre-process the data.
These steps include:
\begin{itemize}
  \item Normalizing the various forms of the letter "alif" \includegraphics[scale=0.2]{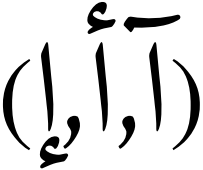}
   to the letter \includegraphics[scale=0.2]{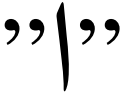}
  .
  \item Removing all diacritic marks.  
\end{itemize}

\begin{table}[t]
\centering
\begin{tabular}{|l|l|}
\hline
 Original text & \includegraphics[scale=0.2]{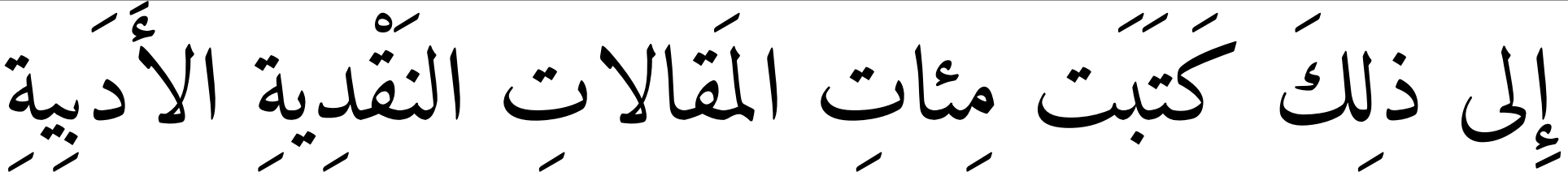}
 \\[1.5ex] \hline 
 
 pre-processed text& \includegraphics[scale=0.2]{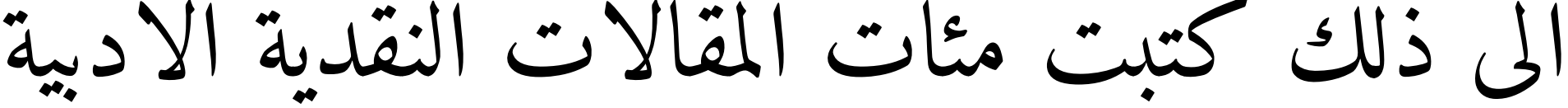}
 \\[1.5ex] \hline
\end{tabular}
\caption{An example on how we pre-process Arabic text. The letter "alif" is normalized and all diacritic marks are removed.}
\label{tab:preprocess_example}
\end{table}

We show an example of an original and  pre-processed sentence from OntoNotes 5.0 in Table \ref{tab:preprocess_example}. Pre-processing the data increases the overall performance of coreference system with 7 percentage points more as we will see in Section \ref{sec:results}.

\subsection{Multilingual vs. monolingual BERT}
\label{sec:bert}
{\BERT}  \cite{bert:2019} is a language representation model consisting of multiple stacked Transformers \cite{vaswani-et-al:2017:attention}. 
{\BERT} was pretrained on a large amount of unlabeled text,
and produces distributional vectors for words and contexts. Recently, it has been shown that {\BERT} can capture structural properties of a language, such as its surface, semantic, and syntactic aspects \cite{jawaher:2019} which seems related to what we need for the coreference resolution. Therefore, we set {\BERT} to produce embeddings for the mentions. {\BERT} is available for English, Chinese, and there is a version for multiple languages, called multilingual {\BERT} \footnote{https://github.com/google-research/bert}. Multilingual {\BERT} is publicly available and covers a wide range of languages including Arabic. Even though the multilingual version provides great results for many languages, it has been shown their monolingual counterparts to achieve better. 
Therefore, recent research adopts the monolingual approach to pretrain {\BERT}, developing, e.g., \ACRO{CamemBERT} for French \cite{martin:2019}, \ACRO{AlBERTo} for Italian \cite{polignano:2019}, and others \cite{Leekr:2020,Souza:2019,Kuratov:2019}. 
{\ARABERT} \cite{Antoun:2020} is  a monolingual {\BERT} model for Arabic which was pre-trained on a collection of Wikipedia and newspaper articles. There are two versions, {\ARABERT} 0.1 and {\ARABERT} 1.0,
the difference being that the latter pretrained  on the word morphemes obtained using Farasa \cite{darwish2016farasa}. The two versions yield relatively similar scores in various {\NLP} tasks. In our experiments, we used {\ARABERT} 0.1 because empirically it proved  more compatible with the coreference resolution system. 

\subsection{Mention Detection}
\label{sec:mention_detector}
Mention detection is a crucial part of the coreference resolution system, better candidate mentions usually lead to better overall performance. As suggested by \newcite{yu-etal-2020-neural}, a separately trained mention detector can achieve a better mention detection performance when compared to its end-to-end counterpart. In this work, we adapt the state-of-the-art mention detector of \newcite{yu-etal-2020-neural} to aid our system. In their paper, \newcite{yu-etal-2020-neural} evaluated three different architectures for English mention detection task,  we use their best settings (\textsc{biaffine md}) and replace their {\ELMO} embeddings with {\BERT} embeddings in the same way we did for our coreference system\footnote{We tried to add the fastText and character-based embeddings to the system but found they do not improve the mention detection results}. The \textsc{biaffine md} uses contextual word embeddings and a multi-layer bi-directional LSTM to encode the tokens. It then uses a biaffine classifier \cite{dozat-and-manning2017-parser} to assign every possible span in the sentence a score. Finally, the candidate mentions are chosen according to their scores. In addition to the standard high-F1 setting, the system has a further option (high-recall) to output top mentions in the proportion of the number of tokens, this is similar to our mention detection part of the system. Here we use the high-recall settings of the mention detector we modify the baseline system to allow the system using the mentions supplied by the external mention detector. 

\begin{algorithm}[t]
\SetAlgoLined
\KwInput{Training step: $N$; Candidate mentions from external mention detector: $\textsc{Candidate}_{\textsc{external}}$}
\KwOutput{Trainable variables: $W$}
$n=0$\;
\While{$n \leq N$ }{
$ \textsc{pipeline}_{\textsc{ratio}} \leftarrow n/N $\;
$ rand = random.random()$\;
\uIf{$rand \leq \textsc{pipeline}_{\textsc{ratio}}$}{
$\textsc{CandidateMention} \leftarrow \textsc{Candidate}_{\textsc{external}}$\;
}\Else{
Generate mention candidates $\textsc{Candidate}_{\textsc{end-to-end}}$\;
$\textsc{CandidateMention} \leftarrow \textsc{Candidate}_{\textsc{end-to-end}}$\;
}
Predict antecedent for candidate mentions\;
Compute training loss\;
Update $W$\;
$n \leftarrow n+1$
}
\caption{End-to-end annealing algorithm.}\label{algorithm}
\end{algorithm}

To confirm our hypothesis that a separately trained mention detector can achieve a better mention detection performance, we compare the mention detection performance of our system with the separately trained mention detector. For our system, we train the models end-to-end and assess the quality of candidate mentions before feeding them into the mention clustering part of the system. Table \ref{tab:mention_detector} shows the comparison of both systems in three different settings (\textsc{multiBert} (baseline), \textsc{multiBert+pre} (multilingual {\BERT} and data pre-processing), \textsc{araBert+pre} ({\ARABERT} and data pre-processing)). As we can see from the table, the separately trained mention detector constantly have a better recall of up to 3\% when compared with the jointly trained mention detector\footnote{Here we only care about the recall as the number of candidate mentions is fixed}.

\begin{table}[t]
\centering
\begin{tabular}{lcccccc}
\toprule
 \multirow{2}{*}{Models}&  \multicolumn{3}{c}{Joint}  & \multicolumn{3}{c}{Separate}\\ \cmidrule{2-7}
&R&P&F1&R&P&F1\\ \midrule
\textsc{baseline (multiBert)}&85.6	&24.4	&38.0&\bf 88.1	&25.2	&39.2\\
\textsc{multiBert+pre}&91.2	&26.0	&40.5&\bf 93.3	&26.6	&41.5\\
\textsc{araBert+pre}&92.5	&26.4	&41.1&\bf 95.5	&27.2	&42.4\\
\bottomrule
\end{tabular}
\caption{The mention detection performance comparison between the separately and jointly trained mention detectors in a high recall setting.}
\label{tab:mention_detector}
\end{table}

The preliminary experiments show that by simply using the mentions generated by the external mention detector in a pipeline setting result in a lower coreference resolution performance. We believe this is mainly because in an end-to-end setting, the model is exposed to different negative mention examples; hence, has a better ability to handle false positive candidates. To leverage the benefits between better candidate mentions and more negative mention examples, we introduce a new hybrid training strategy (end-to-end annealing) that initially training the system in an end-to-end fashion and linearly decreasing the usage of end-to-end approach. At the end of the training, the system is trained purely in a pipeline fashion. The resulted system is then tested in a pipeline fashion. Algorithm \ref{algorithm} shows the details of our end-to-end annealing training strategy.

\section{Experimental Setup}
\label{sec:evaluation}
Since the {\BERT} models are large, 
the fine-tuning approaches are more computationally expensive: GPU/TPUs with large memory (32GB+) are required. 
In this work, we use {\BERT} embeddings in a pre-trained fashion to make our experiment feasible on a GTX-1080Ti GPU with 11GB memory.
\subsection{Dataset}
We run our model on the Arabic portion of  OntoNotes 5.0, which were used in the  the official {\CONLL}-2012 shared task \cite{conll:2012}. The data is divided into three splits: train, development, and test. We used each split for its purpose, the train for training the model, the development for optimizing the settings, and the test for evaluating the overall performance. Detailed information about the number of documents, sentences, and words can be found in Table \ref{tab:ontoinfo}.
\begin{table}[t]
\centering
\begin{tabular}{lccc}
\toprule
 Category & Training & Dev & Test    \\ \midrule
 Documents & 359 & 44 & 44 \\
 Sentences & 7,422 & 950 & 1,003  \\
 Words & 264,589 & 30,942 & 30,935 \\
\bottomrule
\end{tabular}
\caption{Statistics on Arabic portion of {\CONLL}-2012.}
\label{tab:ontoinfo}
\end{table}
\subsection{Evaluation Metrics}
For our evaluation on the coreference system, we use the official {\CONLL} 2012 scoring script v8.01 to score our predictions.
Following standard practice, we report recall, precision, and F1 scores for MUC, B$^3$ and CEAF$_{\phi_4}$ and the average F1 score of those three metrics. For our experiments on the mention detection we report recall, precision and F1 scores for mentions.

\subsection{Hyperparameters}
We use the default settings of \newcite{lee:2018}, and replace their GloVe/{\ELMO} embeddings with the fastText/{\BERT} embeddings. Table \ref{tab:config} shows the hyperparameters used in our system.

\begin{table}[t]
    \centering
    \begin{tabular}{l l}
    \toprule
    \bf Parameter & \bf Value \\
    \midrule
    bi-directional LSTM layers/size/dropout &3/200/0.4\\
    FFNN layers/size/dropout & 2/150/0.2\\
    CNN filter widths/size& [3,4,5]/50\\
    Char/fastText/Feature embedding size&8/300/20\\
    {\BERT} embedding size/layer&768/Last 4\\
    Embedding dropout & 0.5\\
    Max span width &30\\
    Max num of antecedents&50\\
    Mention/token ratio &0.4\\
    Optimiser & Adam (1e-3)\\
    Training step & 400K\\
    \bottomrule
    \end{tabular}
    \caption{Hyperparameters for our models.}
    \label{tab:config}
\end{table}

\section{Evaluation}
\label{sec:results}
\subsection{Results}
\textbf{Baseline} We first evaluate our baseline system using the un-pre-processed data and the multilingual {\BERT} model. As we can see from Table \ref{tab:results_coref}, the baseline system already outperforms the previous state-of-the-art system which is based on handcrafted features by a large margin of 2.6 percentage points. The better F1 scores are mainly as a result of a much better precision in all three metrics evaluated, the recall is lower than the previous state-of-the-art system \cite{anders:2014}. 

\textbf{Data pre-processing} We then apply heuristic rules to pre-process the data. The goal of pre-processing is to reduce the sparsity of the data by normalizing the letters that have different forms and removing the diacritics. By doing so, we created a 'clean' version of the data. As we can see from Table \ref{sec:results}, the simple pre-processing on the data achieved a large gain of 7 percentage points when compared with the baseline model trained on the original data. Since the pre-processing largely reduced the data sparsity, the recall of all three matrices has been largely improved. We further compare the mention scores of two models (see Table \ref{tab:results_mention}). As illustrated in the table, the system trained on the pre-processed data achieved a much better recall and a similar precision when compared with the baseline. This suggests that data pre-processing is an efficient and effective way to improve the performance of the Arabic coreference resolution task.

\textbf{Language Specific BERT Embeddings} Next, we evaluate the effect of the language-specific {\BERT} embeddings. The monolingual {\BERT} model ({\ARABERT}) trained specifically on Arabic Wikipedia and several news corpora has been shown that it can outperform the multilingual {\BERT} model on several {\NLP} tasks for Arabic. Here we replace the multilingual {\BERT} model with the {\ARABERT} model to generate the pre-trained word embeddings. We test our system with {\ARABERT} on the pre-processed text, the results are shown in Table \ref{tab:results_coref} and Table \ref{tab:results_mention}. As we can see from the Tables, the model enhanced by the {\ARABERT} achieved large gains of 4.7 and 2.4 percentage points when compared to the model using multilingual {\BERT} on coreference resolution and mention detection respectively. Both recall and precision are improved for all the metrics evaluated which confirmed the finding in \newcite{Antoun:2020} that  {\ARABERT} model is better suited for Arabic {\NLP} tasks.

\begin{table}[t]
\centering
\resizebox{\textwidth}!{
\begin{tabular}{lcccccccccc}
\toprule
\multirow{2}{*}{Models} & \multicolumn{3}{c}{MUC} & \multicolumn{3}{c}{B$^{3}$} & \multicolumn{3}{c}{CEAF$_{\phi_4}$} & Avg. \\ \cmidrule{2-10} 
& R & P & F1 & R & P & F1 & R & P & F1 & F1\\ \midrule
\newcite{bjorkelund:2011}&43.9 &52.5& 47.8& 35.7 &49.8 &41.6 & 40.5& 41.9 &41.2 &43.5\\
\newcite{fernandes-etal-2012-latent}  & 43.6	&49.7	&46.5	&38.4	&47.7	&42.5	&48.2	&45.0	&46.5	&45.2\\
\newcite{anders:2014}  & 47.5	&53.3	&50.3	&44.1	&49.3	&46.6	&49.2	&49.5	&49.3	&48.7 \\ \midrule
 
\textsc{baseline (multiBert)} & 45.7	&66.9	&54.3	&38.8	&64.3	&48.4	&45.7	&57.9	&51.1	&51.3\\
\textsc{multiBert+pre}&56.1	&67.1	&61.1	&50.0	&63.4	&56.0	&54.8	&61.1	&57.8	&58.3\\
\textsc{araBert+pre}&62.3	&70.8	&66.3	&56.3	&65.8	&60.7	&58.8	&\bf 66.1	&62.2	&63.1\\
\textsc{araBert+pre+md}&\bf 63.2	&\bf 70.9	&\bf 66.8	&\bf 57.1	&\bf 66.3	&\bf 61.3	&\bf 61.6	&65.5	&\bf 63.5	&\bf 63.9\\
\bottomrule
\end{tabular}
}
\caption{Coreference resolution results on Arabic test set.}
\label{tab:results_coref}
\end{table}

\begin{table}[t]
\centering
\begin{tabular}{lccc}
\toprule
 Models & R & P & F1    \\ \midrule
\textsc{baseline (multiBert)}& 56.5	&79.1	&65.9\\
\textsc{multiBert+pre}&67.4	&78.8	&72.6\\
\textsc{araBert+pre}&70.6	&79.9	&75.0\\
\textsc{araBert+pre+md}&\bf 72.9	&\bf 80.4	&\bf 76.4\\
\bottomrule
\end{tabular}
\caption{Mention detection results on Arabic test set.}
\label{tab:results_mention}
\end{table}

\textbf{External Mention Detector} Finally, we use a separately trained mention detector to guide our models with a better candidate mentions. We train a mention detector using the same {\CONLL} 2012 Arabic datasets and store the top-ranked mentions in the file. We use the top-ranked mentions from the external mention detector in a pipeline fashion, the mentions are fixed during the training of the coreference resolution task.  We use the output of the mention detector model trained on the pre-processed data and using the {\ARABERT} embeddings as this model performs best over three settings we tested (see Table \ref{tab:mention_detector}). We use the end-to-end annealing training strategy proposed in Section \ref{sec:mention_detector} to train our model with both end-to-end and pipeline approaches. The model is then tested in a pipeline fashion. Table \ref{tab:results_coref} shows our results on coreference resolution, the model enhanced by the external mention detector achieved a gain of 0.8\% when compared to the pure end-to-end model. We further compared the mention detection performance between two models in Table \ref{tab:results_mention}, as expected the new model has a much better mention recall (2.3\%) when compared to the pure end-to-end model (\textsc{araBert+pre}), this suggests our training strategy successfully transferred the higher recall achieved by the external mention detector to our coreference system. 

Overall, our best model enhanced by the data pre-processing, monolingual Arabic {\BERT} and the external mention detector achieved a {\CONLL} F1 score of 63.9\% and this is 15.2 percentage points better than the previous state-of-the-art system \cite{anders:2014} on Arabic coreference resolution.

\subsection{Discussion}
Coreference resolution is a difficult task, 
and even more so for languages such as Arabic 
with more limited resourced.
The main challenge is the lack of large scale coreference resolution corpora. 
At present there are two multilingual coreference corpora that cover Arabic. 
The first is the Automatic Content Extraction (ACE) \cite{doddington2004automatic} which has \textasciitilde500,000 tokens, but mentions are restricted  to  seven  semantic  types\footnote{The semantic types are person, organization, geo-political entity, location, facility, vehicle, and weapon.} and some can be singletons (mentions that do not corefer). 
The second is OntoNotes \cite{conll:2012},
which covers 
all 
entities and does not consider singletons, but the size is smaller than ACE, with \textasciitilde300,000 tokens. A summary of the two corpora in Table \ref{tab:arabic_corpora}. OntoNotes has been the standard for coreference resolution evaluation since the  {\CONLL}-2012 shared task. 
However, its Arabic portion is small and this scarcity poses a considerable barrier to improving coreference resolution. 

Another challenge of the task is the absence of large pre-trained language  models. There are two versions of {\BERT}: {\BERT}-base and {\BERT}-large. 
{\BERT}-large integrates more parameters to encode better representations for mentions which usually leads to a better performance in many {\NLP} tasks. {\ARABERT} and multilingual {\BERT} are pre-trained using the {\BERT}-base approach because {\BERT}-large is computationally expensive. We are not aware of any publicly available {\BERT}-large for Arabic that we could have used in our experiments. We surmise that a {\BERT}-large version of Arabic can improve the overall performance as shown in prior works  \cite{joshi2019bert,kantor2019coreference}. 

\begin{table}[t]
\centering
\begin{tabular}{lccc}
\toprule
 Corpora & Language & Tokens & Documents    \\ \midrule
            \multirow{4}{*}{ACE} & \multicolumn{1}{l}{English} & \multicolumn{1}{l}{\textasciitilde960,000} & - \\
                                 & \multicolumn{1}{l}{Chinese} & \multicolumn{1}{l}{\textasciitilde615,000} & -\\
                                 & \multicolumn{1}{l}{Arabic} & \multicolumn{1}{l}{\textasciitilde500,000} & - \\ \hline
                    \multirow{4}{*}{OntoNotes} & \multicolumn{1}{l}{English} & \multicolumn{1}{l}{\textasciitilde1,600,000} & 2384 \\
                                 & \multicolumn{1}{l}{Chinese} & \multicolumn{1}{l}{\textasciitilde950,000} & 1729\\
                                 & \multicolumn{1}{l}{Arabic} & \multicolumn{1}{l}{\textasciitilde300,000} & 447\\

\bottomrule
\end{tabular}
\caption{General domain coreference resolution corpora that include Arabic.}
\label{tab:arabic_corpora}
\end{table}


\section{Conclusion}
\label{sec:conclusion}
In this paper, we modernize the Arabic coreference resolution task by adapting state-of-the-art English coreference system to the Arabic language. We start with a strong baseline system and introduce three methods (data pre-processing, language-specific {\BERT}, external mention detector) to effectively enhance the performance of the Arabic coreference resolution. Our final system enhanced by all three methods achieved a {\CONLL} F1 score of 63.9\% and improved the state-of-the-art result on Arabic coreference resolution task by more than 15 percentage points. 


\section*{Acknowledgements}
This research was supported in part by  the DALI project, ERC Grant 695662, in part by the Human Rights in the Era of Big Data and Technology (HRBDT) project,  ESRC grant ES/M010236/1.

\bibliographystyle{coling}
\bibliography{coling2020}

\end{document}